\documentclass[11pt,a4paper]{article}
\usepackage{times,latexsym}
  
\usepackage{url}
\usepackage[T1]{fontenc}
\usepackage{flushend}
\usepackage{balance}

\newcommand{\squishlist}{
 \begin{list}{$\bullet$}
  { \setlength{\itemsep}{0pt}
     \setlength{\parsep}{1pt}
     \setlength{\topsep}{1pt}
     \setlength{\partopsep}{0pt}
     \setlength{\leftmargin}{1.5em}
     \setlength{\labelwidth}{1em}
     \setlength{\labelsep}{0.5em} } }
 \newcommand{\squishend}{\end{list}}

\usepackage[acceptedWithA]{tacl2018v2}
\usepackage[]{tacl2018v2}
\usepackage{xspace,mfirstuc,tabulary}

\usepackage{svg}
\usepackage{csquotes}
\usepackage{amsmath}
\usepackage{multirow}
\usepackage{adjustbox}
\usepackage{booktabs}
\usepackage{tabularx,lipsum,environ,amsmath,amssymb}
\usepackage{microtype}
\definecolor{mygreen}{rgb}{0, 0.5, 0}

\usepackage{comment}
\usepackage[multiple]{footmisc}

\title{Predicting Document Coverage for Relation Extraction}

\author{Sneha Singhania, Simon Razniewski, Gerhard Weikum \\
 Max Planck Institute for Informatics, Germany \\
  {\sf \tt\{ssinghan, srazniew, weikum\}@mpi-inf.mpg.de} \\
}

\date{}

\begin{document}

\makeatletter
\newcommand{\problemtitle}[1]{\gdef\@problemtitle{#1}}
\newcommand{\probleminput}[1]{\gdef\@probleminput{#1}}
\newcommand{\problemquestion}[1]{\gdef\@problemquestion{#1}}
\newcommand{\problemoutput}[1]{\gdef\@problemoutput{#1}}
\NewEnviron{problem}{
  \problemtitle{}\probleminput{}\problemquestion{}
  \BODY
  \par\addvspace{.5\baselineskip}
  \noindent
  \begin{tabularx}{\linewidth}{@{\hspace{\parindent}} lX}
    \multicolumn{2}{@{\hspace{\parindent}}l}{\@problemtitle} \\
    \textbf{Input:} & \@probleminput \\
    \textbf{Question:} & \@problemquestion \\
    \textbf{Output:} & \@problemoutput \\
  \end{tabularx}
  \par\addvspace{.5\baselineskip}
}
\makeatother

\maketitle

\begin{abstract}

This paper presents a new task of predicting the coverage of a text document for relation extraction (RE): does the document contain
many relational tuples for a given entity?
Coverage predictions are useful in selecting the best documents for knowledge base construction with large input corpora. 
To study this problem, we present a dataset of $31,366$ diverse documents for $520$ entities. 
We analyze the correlation of document coverage with features like length, entity mention frequency, Alexa rank, language complexity and information retrieval scores.
Each of these features has only moderate predictive power.
We employ methods combining features with statistical models like TF-IDF and language models like BERT. The model combining features and BERT, HERB, achieves an  F1 score of up to $46\%$. We demonstrate the utility of coverage predictions on two use cases: KB construction and claim refutation.
\end{abstract}

\section{Introduction}

\noindent{\bf Motivation and Problem}
Relation extraction (RE) from text documents is an important NLP task with a range of downstream applications \cite{Han2020}.
For these applications, it is vital to understand the quality of RE results.
While extractors typically provide confidence (or precision) scores, this paper puts forward the notion of \emph{RE coverage} (or recall). 
Given an input document and an RE method, coverage measures the fraction of the extracted relations compared to the complete ground-truth that holds in reality. 
We consider this on a per-subject and per-predicate basis, for example, all memberships of Bill Gates in organizations or all companies founded by Elon Musk.

\begin{figure}[t]
\centering
\includegraphics[width=\linewidth, 
trim={7.5cm 3.7cm 7.5cm 0.5cm},
clip=true]{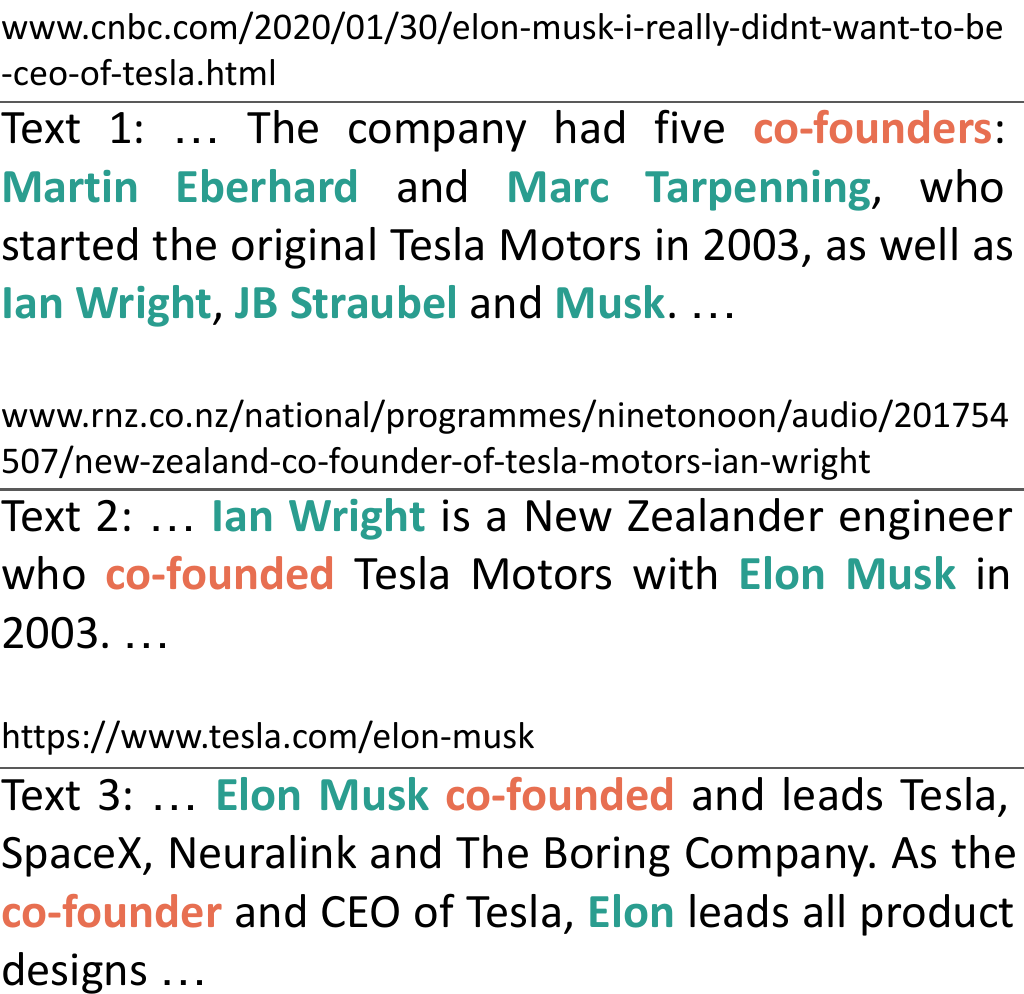}
\caption{Sample documents from our \underline{Do}cument \underline{Co}verage (DoCo) dataset.
}
\label{fig:tesla}
\end{figure}

Document coverage for RE highly varies.
Consider the three text snippets about Tesla as shown in Figure~\ref{fig:tesla}. 
The first text contains all five founders of Tesla, while the second text contains only two of them, and the third has just one. 
Analogously, for the entity Tesla and the relation \textit{founded-by}, we see that text $1$ has coverage $1$, text $2$ has coverage $0.4$, and text $3$ has coverage $0.2$.

\begin{table*}[t]
\adjustbox{max width=\textwidth}{%
\centering
\begin{tabular}{ccp{14cm}c}
 \toprule
 \textbf{Entity} & \textbf{Relation} & \multicolumn{1}{c}{\textbf{Content}} & {\textbf{Coverage}}\\
\toprule
\multirow{4}{*}{George W. Bush} & \multirow{4}{*}{\textit{family}} & \small{President Bush grew up in Midland, Texas, as the eldest son of \textcolor{mygreen}{\textbf{Barbara and George H.W. Bush}} ... and met \textcolor{mygreen}{\textbf{Laura Welch}}. They were married in 1977 ... twin daughters: \textcolor{mygreen}{\textbf{Barbara}}, married to \textcolor{mygreen}{\textbf{Craig Coyne}}, and \textcolor{mygreen}{\textbf{Jenna}}, married to \textcolor{mygreen}{\textbf{Henry Hager}}. The Bushes also are the proud grandparents of \textcolor{mygreen}{\textbf{Margaret Laura \enquote{Mila}}}, \textcolor{mygreen}{\textbf{Poppy Louise}}, and \textcolor{mygreen}{\textbf{Henry Harold \enquote{Hal} Hager}} ... } & \multirow{3}{*}{$1$}\\

\multirow{5}{*}{FedEx} & \multirow{5}{*}{\textit{partner-org}} & \small{FedEx Corp. ... to acquire \textcolor{mygreen}{\textbf{ShopRunner}}, the e-commerce ... acquires the International Express business of \textcolor{mygreen}{\textbf{Flying Cargo Group}} ... acquires \textcolor{mygreen}{\textbf{Manton Air-Sea Pty Ltd}}, a leading provider ... acquires \textcolor{mygreen}{\textbf{P2P Mailing Limited}}, a leading ... acquires \textcolor{mygreen}{\textbf{Northwest Research}}, a leader in inventory ... acquires \textcolor{mygreen}{\textbf{TNT Express}} ... acquires \textcolor{mygreen}{\textbf{GENCO}} ... acquires \textcolor{mygreen}{\textbf{Bongo International}} ... acquires the \textcolor{mygreen}{\textbf{Supaswift businesses}} in South Africa ... acquires \textcolor{mygreen}{\textbf{Rapidão Cometa}} ... } & \multirow{3}{*}{$1$}\\

\multirow{4}{*}{Warren Buffett} & \multirow{4}{*}{\textit{member-of}} & \small{He formed \textcolor{mygreen}{\textbf{Buffett Partnership Ltd.}} in 1956, and by 1965 he had assumed control of \textcolor{mygreen}{\textbf{Berkshire Hathaway}} ... Following Berkshire Hathaway's significant investment in \textcolor{mygreen}{\textbf{Coca-Cola}}, Buffett became ... director of \textcolor{mygreen}{\textbf{Citigroup Global Markets Holdings}}, \textcolor{mygreen}{\textbf{Graham Holdings Company}} and \textcolor{mygreen}{\textbf{The Gillette Company}} ... } & \multirow{3}{*}{$0.8$}\\

\multirow{3}{*}{Indra Nooyi} & \multirow{3}{*}{\textit{edu-at}} & \small{Nooyi was born in Chennai, India, and moved to the US in 1978 when she entered the \textcolor{mygreen}{\textbf{Yale School of Management}} ... secured her B.S. from \textcolor{mygreen}{\textbf{Madras Christian College}} and her M.B.A. from \textcolor{mygreen}{\textbf{Indian Institute of Management Calcutta}} ...} & \multirow{2}{*}{$0.75$}\\

\multirow{2}{*}{J. K. Rowling} & \multirow{2}{*}{\textit{position-held}} & \small{Rowling is one of the best-selling \textcolor{mygreen}{\textbf{authors}} today ... job of a \textcolor{mygreen}{\textbf{researcher}} and bilingual \textcolor{mygreen}{\textbf{secretary}} for Amnesty International ... position of a \textcolor{mygreen}{\textbf{teacher}} led to her relocating to Portugal ...} & \multirow{2}{*}{$0.67$}\\

\multirow{2}{*}{Apple Inc.} & \multirow{2}{*}{\textit{founded-by}} & \small{ \textcolor{mygreen}{\textbf{Steve Jobs}}, the co-founder of Apple Computers ...  switched over to managing the Apple ``Macintosh'' project that was started ... } & \multirow{2}{*}{$0.33$}\\

\multirow{2}{*}{Intel} & \multirow{2}{*}{\textit{board-member}} & \small{\textcolor{mygreen}{\textbf{Andy D. Bryant}} stepped down as chairman ... \textcolor{mygreen}{\textbf{Dr. Omar Ishrak}} to succeed ... \textcolor{mygreen}{\textbf{Alyssa Henry}} was elected to Intel’s board. Her election marks the seventh new independent director ... } & \multirow{2}{*}{$0.125$}\\

\multirow{2}{*}{3M} & \multirow{2}{*}{\textit{ceo}} & \small{The American multinational conglomerate corporation 3M was formerly known as Minnesota Mining and Manufacturing Company. It’s based in the suburbs ... } & \multirow{2}{*}{$0$}\\
\bottomrule
\end{tabular}}
\caption{Sample entity-relation-document triples for all eight relations present in our DoCo dataset.}
\label{tab:example}
\end{table*}

When applying RE at scale, for example, to populate or augment a knowledge base (KB), an RE system may have to process a huge number of input documents that differ widely in their coverage. As state-of-the-art extractors are based on heavy-duty neural networks \cite{Lin:ACL2016,Zhang:EMNLP2017,Soares:ACL2019,Yao:ACL2019}, processing all documents in a large corpus may be prohibitive and inefficient. Instead, prioritizing the input documents by identifying the best documents with high coverage could be effective. This is why coverage prediction is crucial for large-scale RE, but the problem has not been explored so far.

This problem would be easy if we could first run a neural RE system on each document and then assess the yield, either by comparison to withheld labeled data or by sampling followed by human inspection. However, this is exactly the computational bottleneck that we must avoid. 
The challenge is to estimate document coverage, for a given entity and relation of interest, with inexpensive and lightweight techniques for document processing.

\vspace*{0.2cm}
\noindent{\bf Approach and Contributions}
This paper presents the first systematic approach for analyzing and predicting \emph{document coverage for relation extraction}. To facilitate extensive experimental study on this novel task, we introduce a large \underline{Do}cument \underline{Co}verage (DoCo) dataset of $31,366$ web documents for $520$ distinct entities spanning $8$ relations, along with automated extractions and coverage labels.

We employ a classifier architecture that we call HERB (for \underline{He}u\underline{r}istics with \underline{B}ERT), based on a document's lightweight features and additionally incorporates pretrained language models like BERT without any costly re-training and fine-tuning. The best configuration of this classifier achieves an F1-score of up to $46\%$. The classifier provides scores for its predictions and thus also supports ranking documents by their expected yield for the RE task at hand.

We evaluate our approach against a range of state-of-the-art baselines. Our results show that heuristic features like text length, entity mention frequency, language complexity, Alexa rank, or information retrieval scores have only moderate predictive power. However, in combination with pre-trained language models, the proposed classifier gives useful predictions of document coverage.

We further study the role of coverage prediction in two extrinsic use cases: \emph{KB construction} and \emph{claim refutation}. For KB construction, we show that coverage estimates by HERB are effective in ranking candidate documents and can substantially reduce the number of web pages one needs to process for building a reasonably complete KB. For the task of claim refutation (e.g., Tim Cook is the CEO of Microsoft), we show that coverage estimates for different documents can provide counter-evidence that can help to invalidate
false statements obtained by RE systems.

\noindent
The salient contributions of this work are:
\squishlist
\item[1.] We introduce the novel task of predicting document information coverage
for RE.

\item[2. ] To support experimental comparisons, we present a large dataset of annotated web documents. 
\item[3.] We propose a set of heuristics for coverage estimation, analyze them in isolation and in combination with an inexpensive standard embedding-based document model.  

\item[4.] We study the application of our classifier on two important use cases: KB construction and claim refutation.
Experiments show that our predictor is useful in both of these tasks.
\squishend

\noindent
Our data, models and code is publicly available\footnote{\url{www.mpi-inf.mpg.de/document-coverage-prediction}}.

\section{Related Work}

\paragraph{Relation Extraction (RE)}
RE is the task of identifying the relation types between two entities that are mentioned together in a sentence or in proximity within a document (e.g., in the same paragraph).
RE has a long history in NLP research \cite{Mintz:ACL2009,Riedel:ECML2010}, with a recent overview given by \citet{Han2020}. State-of-the-art methods are based on deep neural networks trained via distant supervision \cite{Lin:ACL2016,Zhang:EMNLP2017,Soares:ACL2019,Yao:ACL2019}. On the practical side, RE is available in several commercial APIs for information extraction from text. In our experiments, we make use of Rosette{\footnote{\url{https://rosette.com/}}} and Diffbot{\footnote{\url{https://www.diffbot.com/}}}. Our approach is agnostic to the choice of extractors, though; any RE tool can be plugged in.

\paragraph{Knowledge Base Construction (KBC)}
RE plays a crucial part in the more comprehensive KBC task: identifying instances of entity pairs that stand in a given relation in order to construct a knowledge base \cite{mitchell2018never, Weikum2020MachineKC, Hogan2021KnowledgeG}.

The input is typically a set of documents, often assumed to be fixed and given upfront. This disregards the critical issue of benefit/cost trade-offs, which mandates identifying high-yield inputs for resource-bounded KBC.
Identifying relevant, expressive and preferable sources for KBC is often referred to as \emph{source discovery}. Source discovery can be performed via IR-style ranking of documents or can be based on heuristic estimators of the yield of relation extractors \cite{wang2019midas,paramita-emnlp-2019}. The former work, in particular, approaches yield optimization as a set coverage maximization problem through shared properties of extracted entities. The latter uses textual features in a supervised SVM or LSTM model, a baseline with which we also compare in our experiments.

\paragraph{Document Ranking in IR}
Information retrieval (IR) ranks documents by relevance to a query with keywords or telegraphic phrases. Relevance judgments are based on the perception of informativeness concerning the query and its underlying user intent. Standard metrics for assessment, like precision, recall and nDCG~\cite{Jrvelin2002CumulatedGE}, are not applicable to 
our setting. The notion of coverage pursued in this paper refers to the yield of structured outputs by RE systems rather than document relevance. For example, a query-topic-wise highly relevant document that contains few extractable facts about named entities would still have low RE coverage.

\paragraph{Relevance of Coverage Estimates}
Understanding and incorporating document coverage prediction into NLP-based information extraction is essential for several reasons.
For \emph{resource-bounded KB construction}, it is crucial to know which documents are most promising for extraction with limited budgets for crawling and RE processing and/or human annotation~\cite{ipeirotis-text-query-optimizer, wang2019midas}. For \emph{claim refutation}, coverage estimates can help to assess statements as questionable if documents with high coverage do not support them. So far, claim evaluation systems mostly rely on textual cues about factuality or source credibility \cite{Nakashole:ACL2014,Rashkin:EMNLP2017, Thorne:NAACL2018,Chen:NAACL2019}.

For \emph{question answering} over knowledge bases, it is important to know whether a KB can be relied upon in terms of complete answer sets \cite{darari2013completeness, hopkinson2018demand,DBLP:journals/corr/abs-2001-04425}. Current coverage estimation techniques for KBs do this analysis only post-hoc after the KB is fully constructed \cite{galarraga2017predicting, luggen2019non}, losing access to valuable information from extraction time.

\section{Coverage Prediction}\label{sec:heuristics}

We take an entity-centric perspective, and view RE methods as functions mapping document-entity-relation tuples onto the set of objects found in the document. Formally, given a document $d$, an entity $e$, a relation $r$, a ground truth $\mathtt{GT}$ of objects that stand in relation $r$ with $e$, and a relation extraction method $extr$, the document coverage of $d$ for $(e, r)$ applying $extr$ is defined as:
\begin{equation}
\mathrm{coverage}_{extr}(d,e,r) = \frac{\mid extr(d,e,r) \cap GT \mid }{\mid GT \mid}\label{eq:cov}
\end{equation}
%
The task thus takes the form of a classical prediction problem, either as a numerical coverage value or binarized class label. In Section~\ref{sec:approach}, we propose several heuristics and methods that can be used to predict coverage for a given document. To study this novel problem, we require evaluation data. The following section thus deals with the generation of a large and diverse document coverage dataset.

\section{Dataset Construction}\label{sec:data_cons}

\begin{figure*}[htbp]
\centering
\includegraphics[width=\linewidth, trim={0cm 2cm 0cm 0cm}, clip=true]{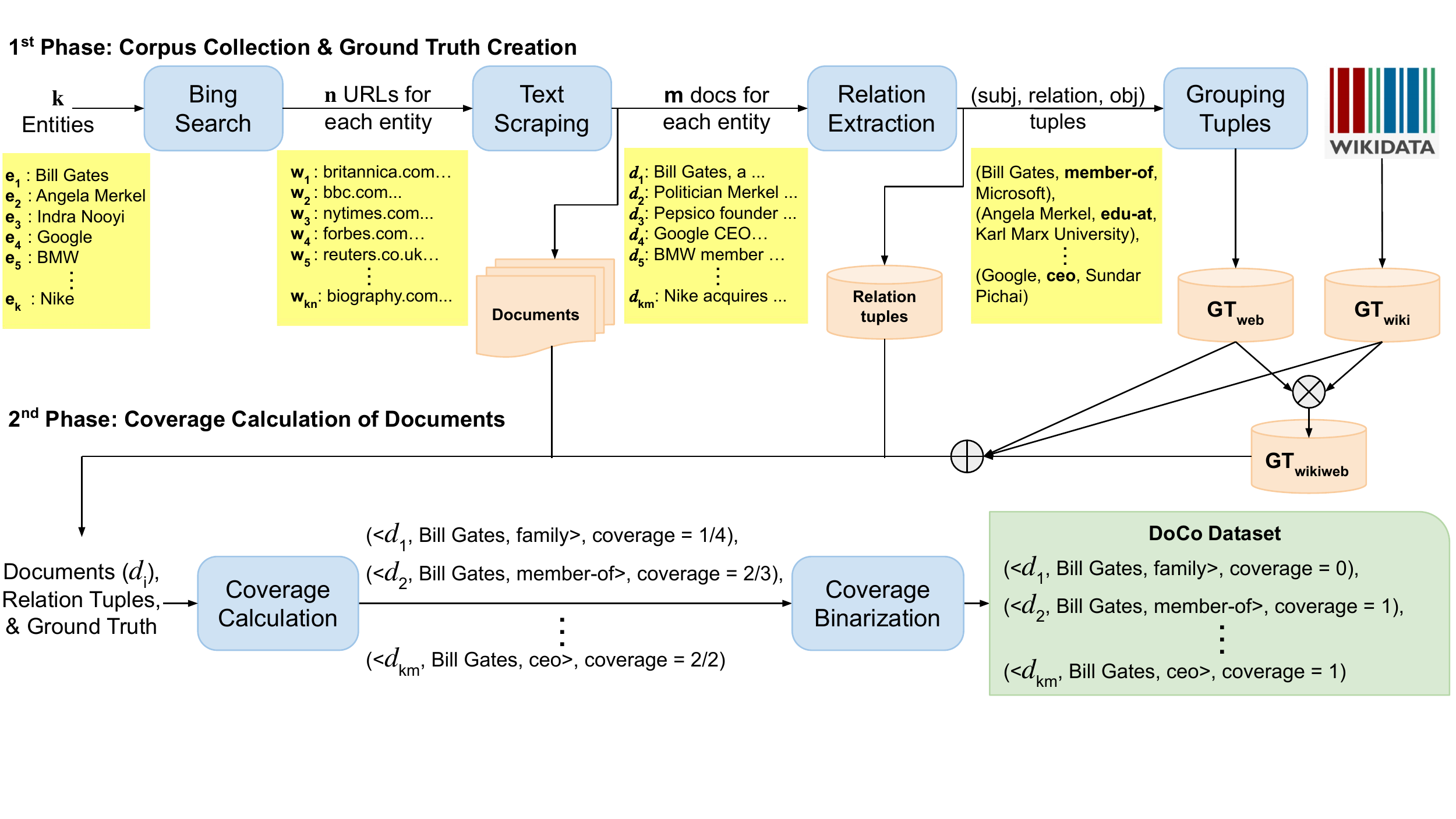}
\caption{\textbf{Dataset Construction Pipeline.} There are two main phases: 1) corpus collection to create $\mathtt{GT_{web}}$, and 2) coverage calculation. Phase 1 involves: i) for each entity $e_i$, $n$ websites are collected using the Bing search API, ii) text is scraped from each website, iii) RE tuples from documents are extracted via Rosette/Diffbot, and iv) RE tuples are deduplicated and consolidated to form $\mathtt{GT_{web}}$. The scraped documents are stored as inputs for phase 2 which consists of: i) for each document $d_i$, previously extracted relations are collected, and ii) based on the choice of $\mathtt{GT}$, coverage is calculated to create the final DoCo dataset.}
\label{fig:inpdata}
\end{figure*}

A thorough study of document coverage prediction requires a corpus with two characteristics: (i) relation diversity, i.e., documents containing enough automatically extractable relations, and (ii) content diversity, i.e., multiple documents with varying content per entity. Existing text corpora, like the popular NYT~\cite{nytimes} and Newsroom dataset~\cite{Grusky2018NewsroomAD}, contain ample numbers of articles that mention newsworthy entities; however, the articles are primarily short, mentioning only very few relations. On the other end, machine-translated multilingual versions of Wikipedia articles~\cite{roy2020topic} allow extraction of many relations but lack diversity.

For the novel task of predicting document information coverage, we thus built the DoCo (\underline{Do}cument \underline{Co}verage) dataset, consisting of $31,366$ web documents for $520$ distinct entities, each with its coverage value. Figure~\ref{fig:inpdata} illustrates the dataset construction.

\paragraph{Entity Selection} First, well-known entities of two types, person (PER) and organization (ORG), were selected from popular ranking lists by Time $100$\footnote{\url{https://time.com/collection/100-most-influential-people-2020/}} and Forbes\footnote{\url{https://forbes.com/forbes-400/}}\footnote{\url{https://forbes.com/lists/global2000/\#9a993675ac04}} (``Influential people around the globe'', ``Most valuable tech companies''). These entities covered $12$ diverse sub-domains, including politicians, entrepreneurs, singers, sportsmen, writers, actors, for PER, and technology, automobile, retail, conglomerate, pharmaceuticals, financial corporations, for ORG. Popular and long-tail entities for PER, companies across demographics and with differing net worth for ORG, were chosen to further obtain documents with varying content. 

\paragraph{Websites \& Content} We aimed to collect diverse $100$ URLs per entity by issuing a set of search engine queries per entity, e.g., ``about PER'', ``PER biography'', ``ORG history''. A total of $6$ set of queries for PER and $10$ for ORG was designed. Since the URLs returned over the set of queries were not always unique, we retained the duplicated URL only once. 

Extracting textual content without noisy headers, menus, and comments, required a labor-intensive scraping step. We handled the multi-domain content scraping task through a combination of libraries like Newspaper3k\footnote{\url{https://newspaper.readthedocs.io/en/latest/}}, Readability\footnote{\url{https://pypi.org/project/readability-lxml/}}, and online scraping services like Import.io\footnote{\url{https://www.import.io/}} and ParseHub\footnote{\url{https://www.parsehub.com/}}. We ensured high-quality scraped content by applying rule-based filters to remove noisy elements like embedded ADs and reference links. The scraped documents covered a range of website domains, including biographical sites, news articles, official company profiles, newsletters, and so on.

\paragraph{Relation Tuples} 
Each document in DoCo was processed by two relation extraction APIs, Rosette and Diffbot. To annotate each document with coverage, we focused only on the entity queried initially to obtain the document. For our experimental study, we selected the following frequently occurring relations: \textit{member-of}, \textit{family}, \textit{edu-at}, and \textit{position-held},  for PER, and \textit{partner-org}, \textit{founded-by}, \textit{ceo}, and \textit{board-member}, for ORG. For more accurate coverage calculation, the RE tuples were deduplicated, e.g., (Gates, \textit{member-of}, Microsoft Corp.) would become (Bill Gates, \textit{member-of}, Microsoft), via alignment to Wikidata identifiers returned by the APIs. 

The relations extracted by the APIs are fine-grained like person-member-of, person-employee-of, org-acquired-by, and org-subsidiary-of. We combined the first two as member-of for PER and the last two as partner-org for ORG as coarse-grained relations.

\begin{table}[t]
\scalebox{0.8}{
\centering
\begin{tabular}{lp{6.5cm}}
\toprule
\textbf{Relation} & \textbf{Wikidata Property}\\
\toprule
\textit{member-of} & member of (P463), member of political party (P102), part of (P361), employer (P108), owner of (P1830), record label (P264), member of sports team (P54)\\
\textit{family} & father (P22), mother (P25), spouse (P26), child (P40), stepparent (P3448), sibling (P3373)\\
\textit{edu-at}  & educated at (P69) \\  
\textit{position-held}  & position held (P39), occupation (P106) \\
\midrule
 \textit{partner-org} & owner of (P1830), owned by (P127), member of (P463), parent organization (P749), subsidiary (P355)\\
\textit{founded-by} & founded by (P112)\\
\textit{ceo} & chief executive officer (P169)\\
\textit{board-member} & board member (P3320)\\
\bottomrule
\end{tabular}}
\caption{Wikidata property names and identifiers used to create $\mathtt{GT_{wiki}}$ }
\label{tab:wikidata_properties}
\end{table}

\paragraph{Ground Truth} We considered three 
ground-truth labels to calculate coverage for each document:

\begin{enumerate}
    \item \textit{Wikidata} ($\mathtt{GT_{wiki}}$): A popular KB providing data for most relations yet having coverage limitations~\cite{galarraga2017predicting, luggen2019non}. For example, for Bill Gates, Microsoft and other popularly associated companies for the \textit{member-of} relation are present, but niche entities like Honeywell are missing. Depending on the entity type and sub-domain, we created the ground-truth labels by choosing those Wikidata properties that best matched the semantics of the $8$ selected relations. Table~\ref{tab:wikidata_properties} provides the complete information. 
    
    \item \textit{Web Extractions} ($\mathtt{GT_{web}}$): We used the set of frequent extractions across all the documents in DoCo as web-aggregated ground truth. For a given entity-relation ($e, r$), an extraction was determined frequent if it appeared in at least $5\%$ of total documents corresponding to $e$, or if its count was no less than $5$ times the highest counted tuple for ($e, r$). Deciding frequent extractions relative to total document count and other tuples' frequencies for an entity resulted in noise-free ground-truth labels.

    \item \textit{Wikidata and Web Extractions} ($\mathtt{GT_{wikiweb}}$): We merged both previous variants using set union operation and phrase embeddings with cosine similarity for higher recall.
\end{enumerate}

\paragraph{Coverage Calculation} Coverage was computed on a per entity-relation-document basis using Equation~\ref{eq:cov}. Even though real-valued coverage values are computed while constructing the dataset, it is often not possible to give nuanced predictions at test time. Consider the text ``... Musk is a co-founder of Tesla ...''. The term \textit{co-founder} clearly indicates the presence of multiple founders; however, the context does not provide any clue on the total number of co-founders. For example, there could be one other co-founder (coverage $0.5$) or $9$ other co-founders (coverage $0.1$). 

\paragraph{Coverage Binarization} We binarized the coverage values to circumvent the above problem, splitting documents into two classes: \emph{informative} and \emph{uninformative}. The binarization method comprised of an absolute and a relative threshold: a document was labeled as informative or $1$ if its coverage was greater than $0.5$, or greater than the coverage of at least $85\%$ of documents for the same $(e,r)$; otherwise, it was labeled as uninformative or $0$. 

\begin{table}[t]
\centering
\scalebox{0.9}{
\begin{tabular}{p{5cm}l}
\toprule
\# PER entities & 250\\
\# ORG entities & 270\\
\# Relations & 8\\
\# Documents & 31,366\\ 
Doc.\ length range (words) & [20, 10906]\\
\# Unique website domains & 600\\
\# Doc. with non-zero RE tuples & 26956\\
\# Doc. with non-zero coverage & 14086\\

\# Doc.\ in class informative & 7103 (22.6 \%)\\
\bottomrule
\end{tabular}
}
\caption{Characteristics of the DoCo Dataset}
\label{tab:datastats}
\end{table}

\begin{table}[t]
\centering
\scalebox{0.9}{
\begin{tabular}{lccc}
\toprule
\textbf{Relation} & $\mathtt{GT_{wiki}}$ & $\mathtt{GT_{web}}$ &  $\mathtt{GT_{wikiweb}}$\\
\toprule
\textit{member-of} & 3.61 & 6.51 & 7.12\\
\textit{family} & 2.21 & 4.0 &  4.41\\
\textit{edu-at} & 2.26 & 2.07 & 2.58\\
\textit{position-held} & 5.86 & 7.76 & 10.37 \\
\midrule
\textit{partner-org} & 6.16 & 4.26 & 3.12\\
\textit{founded-by} & 1.07 & 1.06 & 1.66\\
\textit{ceo} & 1.03 & 2.77 & 2.86\\
\textit{board-member} & 0.47 & 1.44  & 1.75\\
\bottomrule
\end{tabular}
}
\caption{Average number of objects per entity}
\label{tab:datastats2}
\end{table}

\paragraph{Dataset Characteristics}
After filtering duplicates, irrelevant URLs like social media handles, and video-content websites, we obtained a total of $31,366$ documents for $520$ entities. Table~\ref{tab:datastats} provides an overview of the DoCo dataset. We can see that DoCo's labels are imbalanced, as only $22.6\%$ of the documents are informative and $77.4\%$ are uninformative. The count of documents with non-zero RE tuples is higher than those with non-zero coverage since the RE tuples were not always related to the subject entity, hence irrelevant towards coverage calculation.

Table~\ref{tab:datastats2} gives the average number of objects present in each ground truth variant. On average across relations, the number of objects in $\mathtt{GT_{web}}$ is higher than those in $\mathtt{GT_{wiki}}$ by $23.7\%$, and $\mathtt{GT_{wikiweb}}$ is higher than those in $\mathtt{GT_{wiki}}$ by $28.8\%$. This implies that $\mathtt{GT_{web}}$ and $\mathtt{GT_{wiki}}$ can have overlapping objects, and $\mathtt{GT_{web}}$ might contain extra objects towards $\mathtt{GT_{wikiweb}}$ creation.

\setlength{\tabcolsep}{1.3pt}
\begin{table}[tp]
\centering
\scalebox{0.87}{
\begin{tabular}{p{1.3cm}c|cc|ccc}
\toprule
Relation & Human & Diffbot & Rosette & $\mathtt{GT_{wiki}}$ & $\mathtt{GT_{web}}$ &  $\mathtt{GT_{wikiweb}}$\\
\toprule
\textit{member-of} & \multirow{2}{*}{4.36} & \multirow{2}{*}{3.66} & \multirow{2}{*}{\textbf{5.04}} & \multirow{2}{*}{4.54}	& \multirow{2}{*}{7.12}	& \multirow{2}{*}{9.22} \\
\textit{family}  & 4.74	& \textbf{3.82}	& 0.66	& 1.76	& 5.78	& 6.64 \\
\textit{edu-at}  & 1.72 & 2.5 & \textbf{2.52}	& 2.94	& 3.08	& 2.18\\
\textit{position-held} & \multirow{2}{*}{2.9} & \multirow{2}{*}{\textbf{4.26}} & \multirow{2}{*}{-} & \multirow{2}{*}{6.7}	& \multirow{2}{*}{6.14}	& \multirow{2}{*}{9.52}\\
\midrule
\textit{partner-org} & \multirow{2}{*}{3.7}	& \multirow{2}{*}{0.72}	& \multirow{2}{*}{\textbf{2.26}}	& \multirow{2}{*}{0.8} & \multirow{2}{*}{5.04} & \multirow{2}{*}{5.92} \\
\textit{founded-by} & 1.34	& 0.58	& \textbf{1.8}	& 0.78 & 2.84 & 2.96 \\
\textit{ceo} & 2.02	& \textbf{1.96}	& -	& 1.68	& 4.32	& 4.2 \\
\textit{board-member} & \multirow{2}{*}{2.62} & \multirow{2}{*}{\textbf{1.54}} & \multirow{2}{*}{-} & \multirow{2}{*}{2.82}	& \multirow{2}{*}{3.48}	& \multirow{2}{*}{2.64} \\
\bottomrule
\end{tabular}
}
\caption{Average tuple count per relation. The RE tool with higher tuple count (boldfaced) is chosen for each relation.}
\label{tab:data_quality}
\end{table}
\setlength{\tabcolsep}{6pt}

\paragraph{Dataset Quality}
We analyzed the quality of the DoCo dataset by comparing automatic relation extractions to extractions given by human annotators.
A sample of $400$ documents was selected, $50$ per relation, with half from the high-coverage range and the rest from the low-coverage range. Each document was annotated with all correct tuples for the document's main subject entity.

Table~\ref{tab:data_quality} shows the observed averaged counts. We note that the human annotators extracted a substantial number of tuples for all $8$ relations, indicating the richness and breadth of the DoCo documents. The two automatic extractors mostly yielded smaller numbers of tuples, with a few exceptions. These exceptions include spurious tuples, though. The ground-truth variants consistently suggest higher numbers, but except for the conservative $\mathtt{GT_{wiki}}$, these are usually overestimates due to spurious tuples. The $\mathtt{GT}$ variants should thus be seen as upper bounds for the true RE coverage.

We analyzed how well the automatic annotations reflect human annotations' coverage by computing Pearson correlation coefficients for the entire set of $400$ sample documents. For a relation, the RE tool with higher averaged count was chosen for our experiments, and the correlation for (Human, RE) is 0.68. This shows that optimizing for coverage by automatic RE tools is highly correlated with the overarching goal of approximating human-quality outputs. 

\section{Approach}
\label{sec:approach}

We aim to model coverage prediction by processing unstructured document text by inexpensive lightweight techniques. This is crucial for identifying promising documents before embarking on heavy-duty RE.

\paragraph{Heuristics} We devise several simple heuristics involving textual features for document coverage. 
\begin{enumerate}
    \item \textit{Document Length}: The length of a document is a proxy for the amount of information contained. Longer documents may express more relations.
    
    \item \textit{NER Frequency}: Length can be misleading when a document is verbose, yet uninformative. The count of named-entity mentions matching the relation domain (e.g., persons for the relation \textit{family}, or organizations for the relation \textit{member-of}) could correlate with coverage.

    \item \textit{Entity Saliency}: The more frequently an entity is mentioned in a document, the more likely the document expressed relations for that entity.
    
    \item \textit{IR-Relevance Signals}: 
    The surface similarity of the entire document with the input query is another cue.
    We adopt BM25~\cite{Robertson1994OkapiAT}, a classical and still powerful
    IR model for ranking documents, using $\langle e\rangle + \langle r \rangle$ as query, where $e$ and $r$ are the target entity and relation respectively. Recent advances on neural rankers are considered as well
    \cite{Nogueira2019PassageRW}. We follow \citet{Nogueira2020DocumentRW} and use the T5  sequence-to-sequence model \cite{JMLR:v21:20-074}
    to rank documents.  
    
    \item \textit{Website Popularity}: Popular websites may be visited often because they are more informative. We use the standard Alexa rank\footnote{\url{www.alexa.com}} as a measure of popularity.
   
    \item \textit{Text Complexity}: RE methods are effective on simpler text, and may not be able to effectively extract relations from documents written in complex prose. We use the Flesch score \cite{Flesch1949TheAO}, a popular text readability measure. 
    
    \item \textit{Random}: We contrast the predictive power of our proposed methods with two random baselines: a fair coin, and a biased coin maintaining the label imbalance in our test set.
    \end{enumerate}

\paragraph{Methods} We use several inexpensive statistical models for document representation and feed them to a logistic regression classifier.

\begin{enumerate}
    \item[8. ] \textit{Latent Topic Modeling}: Topics in a document could be a useful indicator of coverage. For example, for relation \emph{family}, latent topics like \emph{ancestry} or \emph{personal life} are relevant. We use Latent Dirichlet Allocation (LDA)~\cite{Blei2003LatentDA} to model documents as distributional vectors. 
    
     \item[9. ] \textit{BOW+TFIDF}: A simple yet effective statistic to measure word importance given a document in a corpus is the product of term frequency and inverse document frequency (TF-IDF). We vectorize a document into a Bag-of-Words (BOW) representation with TF-IDF weights.
    
    \item[10. ] \textit{Ngrams+TFIDF}: A document is vectorized using frequent n-grams ($n \leq 3$) with TF-IDF weights.
\end{enumerate}

We employ two neural baselines including LSTM and pre-trained language model (BERT).

\begin{enumerate}
    \item[11. ] \textit{LSTM}: Previous work by \citet{paramita-emnlp-2019} used textual features to estimate the presence of a complete set of objects in a text segment. We adopt their architecture, representing documents using $100$ dimensional GloVe embeddings~\cite{Pennington2014GloveGV}, and processing them in LSTM~\cite{Hochreiter1997LongSM}, followed by a feed-forward layer with ReLU activation before the classifier. 
    
    \item[12. ] \label{item:bert} \textit{Language Model (BERT)}: Without costly re-training or fine-tuning, we utilize pre-trained BERT embeddings~\cite{Devlin2019BERT} in a feature-based approach by extracting activations from the last four hidden layers. As in the original work, these contextual embeddings are fed to a two-layer 768-dimensional BiLSTM before the classifier.
    \end{enumerate}
Our experiments (Sec.~\ref{sec:results}) reveal that each of our proposed heuristics has only a moderate predictive power. We therefore formulate a lightweight classifier to combine heuristics with the best performing statistical model (TF-IDF), or language model (BERT).

\begin{figure}[t]
\centering
\includegraphics[width=0.9\columnwidth, trim={0cm 14.2cm 17.5cm .3cm}, clip=true]{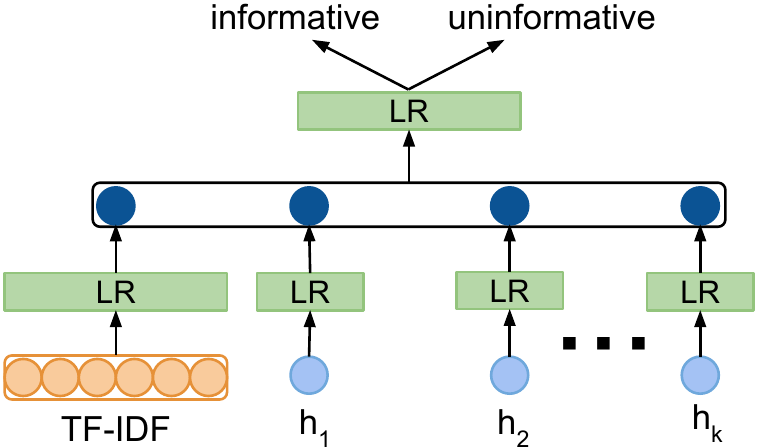}
\caption{Architecture for Heuristics combined with TF-IDF (Heu+TFIDF)}
\label{fig:heu_tfidf}
\end{figure}

\begin{figure}[t]
\centering
\includegraphics[width=0.9\columnwidth, trim={0cm 15.7cm 18.2cm .3cm}, clip=true]{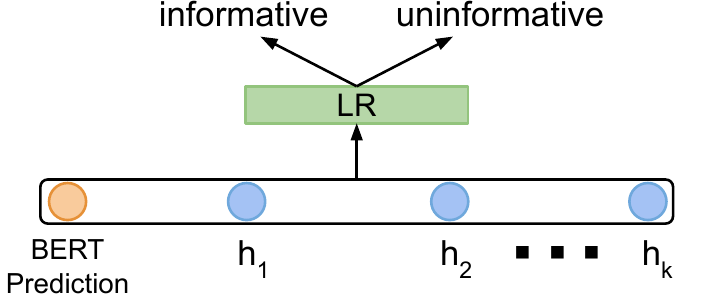}
\caption{Architecture for Heuristics combined with BERT Prediction (HERB)}
\label{fig:hubert}
\end{figure}

\begin{enumerate}
    \item[13. ]\textit{Heuristics with BOW+TFIDF} (Heu+TFIDF): We combine TF-IDF with heuristics (one to six) using stacked Logistic Regression (LR) (Figure~\ref{fig:heu_tfidf}). In level $1$, the TF-IDF vector and each individual heuristic, are fed to separate LR classifiers. In level $2$, all the outputs of level 1 LRs are concatenated and fed to a final LR classifier for coverage prediction. The entire model is jointly trained. 

    \item[14. ]\textit{\textbf{He}u\textbf{r}istics with \textbf{B}ERT} (HERB): We combine BERT with heuristics (one to six) in a two-step process (Figure~\ref{fig:hubert}). In the first step, we reuse the BERT model above (with no additional training or fine-tuning) for coverage prediction. This prediction is then concatenated with heuristics to form a single vector, which is fed to a LR classifier.
\end{enumerate}

\section{Experiments}
\label{sec:experiments}

\subsection{Setup}
\paragraph{Dataset} We considered two automatic RE tools, Rosette and Diffbot, $extr$ (Rosette, Diffbot), and three ground truth variants: $\mathtt{GT_{wiki}}, \mathtt{GT_{web}}, \mathtt{GT_{wikiweb}}$. For each relation, we report on the combination of RE tool and GT variant that achieves the highest count of documents classified as high-coverage.

Each relation had a separate labeled set of documents, split into $70\%$ train, $10\%$ validation and $20\%$ test. Information leakage was prevented by splitting along entities, i.e., all documents on the same entity would exclusively be in one of train, validation or test set. The number of training samples per relation varies from $664$ (\textit{board-member}) to $3604$ (\textit{position-held}). Since the label distribution in DoCo is imbalanced, the uninformative (or $0$) class in all train datasets were undersampled to obtain a $50$:$50$ distribution, while the validation and test datasets were kept unchanged to reflect the real-world imbalance. Named entities and numbers were masked. 

\begin{table*}[htbp]
\centering
\scalebox{0.9}{
\begin{tabular}{p{2.9cm}p{1.2cm}p{1cm}p{1cm}p{1cm}p{1.2cm}p{1.2cm}p{1cm}p{1.4cm}p{.8cm}}
\toprule
& \multicolumn{4}{c}{\textbf{PER}} & \multicolumn{4}{c}{\textbf{ORG}}\\
\cmidrule(lr){2-5} \cmidrule(lr){6-9}
\multirow{2}{*}{\textbf{Method}} & \textit{member-of} & \textit{family} & \textit{edu-at}  &  \textit{position-held} & \textit{partner-org} & \textit{founded-by} & \textit{ceo} &  \textit{board-member} & Avg.\\
\midrule
Random (biased) & 	5.7 & 	6.8 & 	4.9 & 	10.0 & 	7.5 & 	1.2 & 	13.5 & 	3.7 & 6.6\\ 
Random (fair) & 	15.7 & 	11.1 & 	12.6 & 	15.4 & 	15.2 & 	8.9 & 	21.3 & 	7.2 & 13.4\\ 
Text Complexity & 	9.6 & 	5.4 & 	6.1 & 	10.3 & 	3.5 & 	3.3 & 	15 & 	5.4 & 7.3\\ 
Alexa Ranking & 	12.6 & 	9.8 & 	8.1 & 	12.4 & 	16.7 & 	11.3 & 	24.8 & 	7.3 & 12.9\\ 
Entity Saliency & 	17.8 & 	14.3 & 	11.9 & 	18.2 & 	14.7 & 	8.4 & 	24.6 & 	7.1 & 14.6\\ 
Document Length & 	20.5 & 	19.0 & 	15.5 & 	21.9 & 	23.9 & 	12.8 & 	28.8 & 	8.5 & 18.9\\ 
NER Count & 	24.3 & 	19.8 & 	18.2 & 	- & 	21.1 & 	13.7 & 	34.5 & 	11.8 & 20.5 \\ 
BM25 IR & 	27.1 & 	21.1 & 	18.8 & 	26.3 & 	21.8 & 	12.9 & 	36.6 & 	12.1 & 22.1\\ 
T5 IR  &	26.9 &	23.2 &	20.3 &	29.6 &	19.5 &	15.4 &	41.1 &	13.1 & 23.6\\

\midrule
LDA Topic Model & 	19.3 & 	19.0 & 	14.5 & 	21.1 & 	15.7 & 	8.6 & 	25.2 & 	11.5 & 16.9 \\ 
GloVe+LSTM &	16.5 & 	28.6 & 	19.8 & 	32.9 & 	24.2 & 	19.5 & 	24.4 & 	4.9 & 21.3\\ 
Ngrams+TFIDF & 36.2 &	40.0 &	25.6 &	40.2 &	18.6 &	25.5 &	41.8 &	30.2 & 32.3\\
BOW+TFIDF & 	36.0 & 	41.0 & 29.2 & 	42.1 & 	17.2 & 	28.3 & 	40.6 & 	32.1 & 33.3 \\ 
BERT  & 	40.4 & 	39.7 & 	35.7 & 44.4 & 	22.0 & 	30.8 & 43.0 & 33.8 & 36.2\\ 
\midrule
Heu+TFIDF & 41.9 & \textbf{43.5} & 31.3 & 36.5 & \textbf{35.1} & 28.2 & 41.4 & 22.0 & 35.0\\
HERB & \textbf{44.2} & 41.7 & \textbf{40.5} & \textbf{45.6} & 28.8 & \textbf{32.5} & \textbf{46.2} & \textbf{34.8} & \textbf{39.3}\\
\bottomrule
\end{tabular}
}
\caption{\label{tab:intrinsicresults}F1-scores (\%) obtained on the coverage prediction task by various heuristics and methods.}
\end{table*}

\begin{table*}[t]
\centering
\scalebox{0.9}{
\begin{tabular}{p{2.5cm}p{13.5cm}}
\toprule
\textbf{Relation} & \textbf{Important Phrases}\\
\midrule
\textit{member-of} & [org], is part of, ambassador, is associated with, [org] partner\\
\textit{family} & [person], married, father, wife, children, daughter, parents, [number]\\
\textit{edu-at} & [org], graduated, degree, studied, [org] in [number], is part of\\
\textit{position-held} & [person], leader, president, actor, professor, writer, founder, police, portman\\
\textit{partner-org} & [org], [number] [org], subsidiary, merger, the company, member of\\
\textit{founded-by} & [person], founder, director, executive, chairman, co founder, head of, chief executive\\
\textit{ceo} & ceo, [person] director, chief, officer, founders, chief executive officer, president\\
\textit{board-member} & [org], [person], chairman, executive, board of directors, [number] senior executive, officer in charge, representative director\\
\bottomrule
\end{tabular}
}
\caption{\label{tab:topwords} Highly weighted phrases given by the trained LR classifier of Ngrams+TFIDF and BOW+TFIDF.}
\end{table*}

\begin{table}[t]
\centering
\scalebox{0.9}{
\begin{tabular}{p{2.8cm}l}
\toprule

\textbf{HERB} & 39.3\% \\
- Doc. Length &36.8\% (-2.44)\\
- Entity Saliency & 36.4\% (-2.85)\\
- Alexa Ranking & 36.3\% (-3.03)\\
- NER Count & 36.2\% (-3.11)\\
- BM25 & 36.0\% (-3.29)\\
- Text Complexity & 35.7\% (-3.62)\\
\bottomrule
\end{tabular}
}
\caption{Average F1 performance with feature ablations. Text Complexity and BM25 are most important.}
\label{tab:ablation}
\end{table}

\paragraph{Models} Each proposed heuristic was turned into a classifier by first ranking documents according to the heuristic, and then labelling the top $50\%$ documents as class $1$ or informative. We used the Okapi BM25\footnote{\url{https://pypi.org/project/rank-bm25/}} and monoT5\footnote{\url{https://github.com/castorini/pygaggle}} open-source implementations for IR ranking. The monoT5 model is generally used for passage ranking, and as DoCo documents are much longer with multiple passages, we used the MaxP algorithm \cite{Dai2019DeeperTU} to compute the document ranking. Since the difference in performance between T5 and BM25 models is negligible, we chose the simpler yet equally effective BM25 model as IR-relevance signal for HERB.

Feature based methods including topic modeling with LDA, TF-IDF and N-grams, were fed to a Logistic Regression classifier. In the LSTM architecture, we used $100$ dimensional GloVe embeddings with a vocabulary size of $100,000$, and a $100$ dimensional hidden state for LSTM. 

For pre-trained language models, we used the BERT-base-uncased\footnote{\url{https://huggingface.co/bert-base-uncased}} model (without additional retraining or fine-tuning) to encode sentences, by summing the [CLS] token's representation from the last four hidden layers. Input documents were padded or truncated to $650$ sentences, and represented through sentence encodings. Coverage classification was performed using the feature-based approach outlined in ~\citet{Devlin2019BERT}.

We constructed mini-batches of size $32$, used the Adam optimizer initialized with a constant learning rate of $1e$-$05$ and $1e$-$09$ epsilon value, and trained for $200$ epochs. Since our dataset is imbalanced, we monitored validation precision to save the best model, and report optimal F1-scores~\cite{Lipton2014OptimalTO} to compare results.

\begin{table*}[t]
\centering
\scalebox{0.9}{
\begin{tabular}{p{3.7cm}p{1.2cm}p{.9cm}p{1.2cm}p{1.2cm}p{1.2cm}p{1.2cm}p{.8cm}p{1.2cm}p{1cm}}
\toprule

\multirow{2}{*}{\textbf{Setting}} & \textit{member-of} & \textit{family} & \textit{edu-at}  &  \textit{position-held} & \textit{partner-org} & \textit{founded-by} & \textit{ceo} &  \textit{board-member} & \multirow{2}{*}{Avg.} \\
\toprule

HERB (in-domain)	& 40.8 & 41.8 & 34.9 & 42.8 & 28.4 & 17.1 & 45.4 & 23.3 & 34.3\\
HERB (out-of-domain) & 35.7 & 39.7 & 32.5 & 39.1 & 29.4 & 23.8 & 42.3 & 31.1 & 34.2\\
\midrule
Training Data Size & 2194 & 1650 & 1458 & 2940 & 1124 & 828 & 2058 & 608 \\
\bottomrule
\end{tabular}
}
\caption{\label{tab:nodomainmatch} Comparison of F1-scores (\%) of HERB on the in-domain and out--of-domain test set.}
\end{table*}

\subsection{Results} 
\label{sec:results}

Our results are shown in Table~\ref{tab:intrinsicresults}. Each heuristic gives a mediocre performance, with T5 IR achieving the highest average F1 of $23.6$ among the  heuristics. In the trained group of models, LDA has the lowest average F1 of $16.9$, while BERT performs the best with an average F1 of $36.2$.

Although each heuristic has moderate predictive power, combining them with statistical models like TF-IDF, or pre-trained language models like BERT, gives the best performance. Among the combination models, HERB outperforms Heu+TFIDF in a clear majority of relations. 

\paragraph{Model Analysis} Statistical models like BOW+TFIDF and Ngrams+TFIDF performed comparably to BERT for a minority of relations. To better understand these models, we analyzed highly positive and negative features. Table~\ref{tab:topwords} provides noteworthy examples. We observe the presence of semantically relevant phrases. We also inspect the weights of the trained LR classifier of HERB. Across relations, BERT had the highest average weight ($5.05$), followed by BM25 ($2.56$), while NER Count had the lowest weight ($0.07$).

\paragraph{Feature Ablations} We further perform an ablation analysis, with Table~\ref{tab:ablation} showing the average F1-scores when individual heuristics are removed from HERB. Removing either BM25 or Text Complexity leads to a significant drop in performance, indicating that other heuristics or BERT do not capture these features well.

\paragraph{Human Performance} Finally, we compare the results against human performance on identifying high-coverage documents. For each relation, $10$ randomly sampled test documents were labeled as informative or uninformative for RE solely by reading the document. Averaged over all relations, humans obtained an F1 score of $70.42\%$, compared to HERB predictions reaching an average F1 of $39.3\%$, and all baselines were significantly inferior. The large gap between humans and learned predictors shows the hardness of the coverage prediction task and underlines the need for the presented research.

\section{Analysis and Discussion}\label{sec:analysis}
\paragraph{Domain Dependency} To investigate how strongly prediction depends on in-domain training data, we performed a stress test, where the train, validation and test set were split along domains (e.g., singers vs.\ entrepreneurs vs.\ politicians). Table~\ref{tab:nodomainmatch} shows the resulting F1-scores (\%). For HERB, the average F1-score on the in-domain test set is $34.3\%$,  while on the out-of-domain test set is $34.2\%$, i.e., there is no notable drop for the challenging domain-transfer case.  We observe a minor drop for larger relations, while even increases are visible for the smallest two relations. This suggests that HERB learned generalizable features that are beneficial across domains. 

\paragraph{Evaluation of Document Ranking} 
So far, we have evaluated our methods on a binary prediction problem. However, use cases frequently require a ranking capability (see also Sec.~\ref{sec:applications}). We additionally evaluate our methods on a ranking task, where documents are ranked by the score of positive predictions.

We use the mean Normalized Discounted Cumulative Gain (mean nDCG)~\cite{Jrvelin2002CumulatedGE} as the evaluation metric. A similar performance trend to the F1 metric is observed among our methods. HERB performs the best with an average nDCG score of $0.45$ across relations, while BERT and Heu+TFIDF have $0.44$ and $0.43$, respectively.

\paragraph{RE Limitations}
The performance of RE methods significantly impacts the quality of $\mathtt{GT_{Web}}$ as well as the RE coverage of documents. Although we used state-of-the-art commercial APIs, these nonetheless struggle on open web documents. To illustrate this, we randomly sampled $40$ documents from DoCo and compared the count of RE tuples returned by Diffbot/Rosette against the count by a human relation extractor. Diffbot returned $60.6\%$ fewer relational tuples, and Rosette returned $72.3\%$ fewer, suggesting the need for further improvement of RE methods.

\paragraph{Error Analysis}
We analyzed the incorrect predictions by HERB and categorized the errors. For each relation, we randomly sampled 10 incorrectly predicted documents, $5$ false positives and $5$ false negatives. Out of the total $80$ samples, $63.75\%$ of documents contained partial information for the chosen relation; on $15\%$ of documents the IE methods failed to extract all the necessary RE tuples; the ground truth for $3.75\%$ of documents had an incomplete set of objects; $3.75\%$ documents had noisy content; and $2.5\%$ documents had incomplete information due to failure of scraping methods on complex website layouts.

Multiple documents in the low-information category contained speculative content, e.g., considerations about candidates for a new appointment as a board member or CEO. In other cases, the document would mention the increased count of board members, but not their names. A few documents also had partial information leading to false positives, e.g., a document partially talking about the footballer Sergio Agüero for the \textit{family} relation was incorrectly classified as informative; as it also contained a complete family history about another footballer, Diego Maradona (Sergio's father-in-law).

Conversely, documents may contain information relevant to a relation without actual mention of the relation, which leads to false negatives. For example, a document on the LinkedIn Corporation stating `` ... Weiner stepped down from LinkedIn ... He named Ryan Roslansky as his replacement.'' was labeled uninformative for the \textit{ceo} relation. Although Ryan Roslansky and LinkedIn are related through the \textit{ceo} relation, the implicit statement was not noticed by HERB.

We specifically inspected the IR baselines' performance to understand better why these are mediocre predictors at best. The IR signals about entire documents merely reflect that a document is on the proper topic given by the query entity, but that does not necessarily imply that the document contains \textit{many} relational facts about the target entity. For RE coverage, IR-style document-query relevance is a necessary cue but not a sufficient criterion.

\paragraph{Efficiency and Scalability}
We measured the run-time of HERB against a state-of-the-art neural model for document-level RE (DocRED) \hbox{~\cite{Yao:ACL2019}}. Based on the DocRED leaderboard\footnote{\url{https://competitions.codalab.org/competitions/20717\#results}}, we selected the currently best open-source method: the Transformer-based Structured Self-Attention Network (SSAN) \cite{Xu2021EntitySW}.

A sample of $100$ documents from DoCo was given to both HERB and SSAN and processed as follows. For HERB, features are computed utilizing BERT,
followed by coverage prediction. For SSAN, documents first need to be pre-processed to construct the necessary DocRED representation.
This includes named entity recognition and pair-wise co-reference resolution, using Stanza\footnote{\url{https://stanfordnlp.github.io/stanza/}} to properly group same-entity occurrences.

The measurements show the following. HERB takes ca.\ 2 seconds, on average, to process one document, whereas SSAN requires 13.6 seconds -- a factor of 6.8 higher in speed and resource consumption. The difference becomes even more prominent for very long documents with many named entity mentions. HERB's run-time grows linearly with document length, while SSAN's run-time exhibits quadratic growth with the number of entity mentions. 

This quadratic complexity of full-fledged neural RE has inherent reasons (as stated in \citet{Yao:ACL2019}). Document-level relation extraction generally requires computations for all possible pairs of entity mentions.
The neural RE methods need to have the positions of candidate entity pairs as input, which necessitates considering all pairs of mentions.

\section{Applications}\label{sec:applications}

To demonstrate the importance of coverage prediction, we evaluated its utility in two use cases, knowledge base construction and claim refutation. For the former, we discuss the importance of ranking documents by RE coverage (Section~\ref{sec:doc_ranking_re}) and a practically relevant setting where RE is constrained by resource budgets (Section~\ref{sec:budgeted_doc_ranking_re}).

\subsection{Document Ranking for Relation Extraction}
\label{sec:doc_ranking_re}
Relation extraction plays a pivotal role in KB construction. We show the relevance of coverage estimates for prioritizing among documents.  Entities from our test dataset serve as subjects for RE. We select top $k$ documents from the test dataset corpus by four different techniques. We compare the performance of each method by the total number of extracted RE tuples per subject and compute recall w.r.t.\ the Wikidata ground-truth. 

\squishlist
\item[1.] \textit{Random}: A random sample of documents.
\item[2.]  \textit{IR-Relevance}: Using BM25 to identify the most relevant documents.
\item[3.]  \textit{Coverage Prediction}: HERB's predictions to rank documents.
\item[4.]  \textit{Coverage Oracle}: Selecting documents by their ground-truth labels from DoCo. This ranking gives an upper bound on what an ideal method could achieve.
\squishend 

\paragraph{Setup} 
The document coverage calculation is on a per ($e$, $r$) pair basis. In a single iteration, all the proposed methods are given a set of documents partitioned by ($e$, $r$) pairs. Each method uses its technique to rank the documents, and the top $k$ ranked documents are given to the RE API (Rosette or Diffbot) for obtaining the set of relational tuples.

\begin{figure}[t]
\centering
\includegraphics[width=\linewidth]{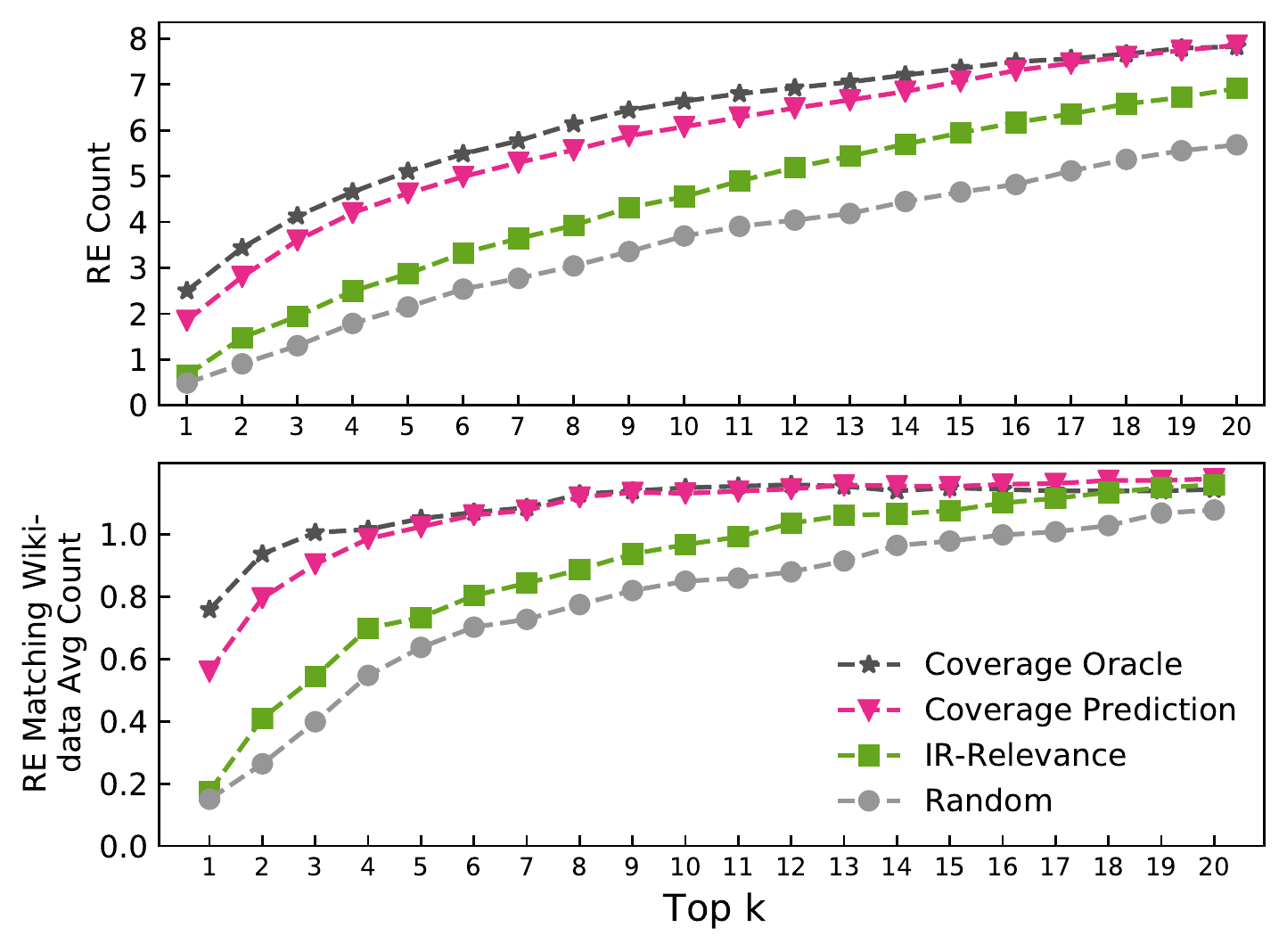}
\caption{Total yield (top) and precision (bottom) of KBC based on different ranking methods for documents.}
\label{fig:total_ie}
\end{figure}

\paragraph{Results} Figure~\ref{fig:total_ie} (\textit{top}) compares the total RE tuples obtained by the proposed methods, averaged across test dataset entities and $8$ chosen relations. Notably, BM25 doesn't perform much better than random, while coverage prediction is not far behind the perfect ranking defined by the coverage oracle. Ordering documents by coverage prediction instead of IR-relevance gives 50\% more extractions from the top-$10$ documents.

Figure~\ref{fig:total_ie} (\textit{bottom}) shows the number of RE tuples that match the Wikidata KB, thus comparing the methods on precision. As was foreseeable, the coverage oracle method wins due to the usage of correct coverage values for ranking. HERB's coverage prediction performance is considerably higher than IR-relevance and other methods, while it matches the coverage oracle for $K \geq 4$. Beyond $K > 15$, all methods yield nearly the same sets of tuples, hence similar precision.

\subsection{Budget-constrained Relation Extraction}
\label{sec:budgeted_doc_ranking_re}

Document coverage predictions are particularly important for massive-scale RE tasks targeted at long-tail entities, such as populating or augmenting a domain-specific knowledge base (e.g., about diabetes or jazz music). Such tasks may require screening a huge number of documents. Therefore, practically viable RE methods need to operate under budget constraints, regarding the monetary cost of computational resources (e.g., using and paying for cloud servers)
as well as the cost of energy consumption and environmental impact. 

In the experiment described here, we simulate this setting, comparing
standard RE by SSAN against HERB-enhanced RE where HERB prioritizes documents for RE by SSAN. We assume a budget of 10 minutes of processing time and give both methods 100 candidate documents. 
SSAN selects documents randomly and processes them until it runs out of time.
HERB+SSAN sorts documents by HERB scores for high coverage and then lets SSAN process them in this order. The time for HERB itself is part of the 10-minute budget for the HERB+SSAN method.

As a proof-of-concept, we ran this experiment for a sample of 10 different entities (each with a pool of 100 documents).

Table~\ref{tab:budgetRE} shows the results. Due to the upfront cost of HERB, HERB+SSAN processes fewer documents within the 10-minute budget, but its yield is substantially higher than that of SSAN alone, by a factor of $1.63$. This demonstrates the need for document-coverage prediction towards realistic usage.

\begin{table}[t]
\centering
\scalebox{0.9}{
\begin{tabular}{p{2.8cm}ll}
\toprule
\textbf{Method} & \textbf{RE Count} & \textbf{\#Docs Processed} \\
\midrule
SSAN & 59 & 410 \\
HERB+SSAN & 96 & 318 \\
\bottomrule
\end{tabular}
}
\caption{Relation extraction under run-time constraint.}
\label{tab:budgetRE}
\end{table}

\begin{table*}[htbp]
\centering
\scalebox{0.9}{
\begin{tabular}{p{2.5cm}p{2.1cm}p{2.2cm}p{8.8cm}}
\toprule
\centering{\textbf{Subject}} & \centering{\textbf{Relation}} &  \centering{\textbf{Object}} & \textbf{Document Snippet}\\
\midrule
\multirow{2}{*}{\small{Alphabet Inc.}} & \centering\multirow{2}{*}{\textit{ceo}} & \multirow{2}{*}{\small{Susan Wojcicki}} & \small{Susan Wojcicki is CEO of Alphabet subsidiary YouTube, which has 2 billion monthly users.}\\
\multirow{2}{*}{\small{Oracle Corporation}} & \centering\multirow{2}{*}{\textit{founded-by}} & \multirow{2}{*}{\small{David Agus}} & \small{Oracle Co-founder Larry Ellison and acclaimed physician and scientist Dr. David Agus formed Sensei Holdings, Inc.}\\
\multirow{2}{*}{\small{PepsiCo}} & \centering\multirow{2}{*}{\textit{board-member}} & \multirow{2}{*}{\small{Joan Crawford}} & \small{Film actress Joan Crawford, after marrying Pepsi-Cola president Alfred N. Steele became a spokesperson for Pepsi.}\\
\bottomrule
\end{tabular}
}
\caption{Incorrect claims extracted by Diffbot RE API from documents predicted as low coverage.}
\label{tab:negclaims}
\end{table*}

\subsection{Claim Refutation}

Our second use case is fact-checking, specifically the case of refuting false claims by providing counter-evidence via RE.

\paragraph{Reasoning}
\textit{Extraction confidence} and \textit{document coverage} are conceptually independent notions. However, when looking at sets of documents, an interesting relation emerges. Consider two documents, $d_1$ with high coverage, and $d_2$ with low coverage, along with two claims $c_1$ and $c_2$ from the respective documents, extracted with the same confidence. \textit{Can we use coverage information to make claims about extraction correctness?}

We propose the following hypothesis: given that $d_1$ is asserted to have high coverage, we can conclude that any statement not mentioned in $d_1$ (like $c_2$) is more likely false. In contrast, the low coverage of $d_2$ implies that $d_2$ is unlikely to contain all factual statements. Thus, $c_1$ not being found in $d_2$ is no indication that it could not be true.

\paragraph{Validation}
We experimentally validated the correctness of the above reasoning as follows. From the collection of relation extractions from the test dataset documents, we randomly sampled 69 pairs of claims for the same entity and relation, which had low support (i.e., extraction found only in one website). We then ordered the pairs by the coverage of the documents that did not express them, obtaining 69 claims with relatively higher coverage in non-expressing documents and 69 claims with relatively lower coverage.

We manually verified the correctness of each claim on the Internet, verifying annotator agreement on a sub-sample, where we found a high Fleiss’ Kappa~\cite{Fleiss1971MeasuringNS} inter-annotator agreement of 0.82.

Using these annotations, we found that from the 69 claims absent from lower-coverage documents, 58\% (40) were correct, while from those absent from higher-coverage documents, only 36\% (25) were correct. In other words, the fraction of correct claims absent from low-coverage documents is $1.6$ times higher; so coverage can be used as a feature for claim refutation.

Table~\ref{tab:negclaims} shows examples of claims absent from high-coverage documents.

\section{Conclusion}

This paper introduces the new task of document coverage prediction and a large dataset for experimental study of the task. Our methods show that heuristic features can boost the performance of pre-trained language models without costly fine-tuning. Moreover, we demonstrate the value of coverage estimates for the use cases of knowledge base construction and claim refutation. Our future research includes developing a user-friendly tool to support knowledge engineers.

\subsection*{Acknowledgments}
We thank Andrew Yates for his suggestions. Further thanks to the anonymous reviewers, action editor, and fellow researchers at MPI, for their comments towards improving our paper. This work is supported by the German Science Foundation (DFG: Deutsche Forschungsgemeinschaft) by grant 4530095897: ``Negative Knowledge at Web Scale''.

\bibliography{tacl}
\bibliographystyle{acl_natbib}

\end{document}